# MMDS: A Multimodal Medical Diagnosis System Integrating Image Analysis and Knowledge-based Departmental Consultation

Yi Ren, HanZhi Zhang, Weibin Li, Jun Fu, Diandong Liu, Tianyi Zhang, Jie He, Licheng Jiao

*Abstract*—We present MMDS, a system capable of recognizing medical images and patient facial details, and providing professional medical diagnoses. The system consists of two core components: The first component is the analysis of medical images and videos. We trained a specialized multimodal medical model capable of interpreting medical images and accurately analyzing patients' facial emotions and facial paralysis conditions. The model achieved an accuracy of 72.59% on the FER2013 facial emotion recognition dataset, with a 91.1% accuracy in recognizing the "happy" emotion. In facial paralysis recognition, the model reached an accuracy of 92%, which is 30% higher than that of GPT-4o. Based on this model, we developed a parser for analyzing facial movement videos of patients with facial paralysis, achieving precise grading of the paralysis severity. In tests on 30 videos of facial paralysis patients, the system demonstrated a grading accuracy of 83.3%. The second component is the generation of professional medical responses. We employed a large language model, integrated with a medical knowledge base, to generate professional diagnoses based on the analysis of medical images or videos. The core innovation lies in our development of a department-specific knowledge base routing management mechanism, in which the large language model categorizes data by medical departments and, during the retrieval process, determines the appropriate knowledge base to query. This significantly improves retrieval accuracy in the RAG (retrieval-augmented generation) process. This mechanism led to an average increase of 4 percentage points in accuracy for various large language models on the MedQA dataset. Our code is open-sourced and available at:
https://github.com/renIIII/MMDS.

## I. INTRODUCTION

Recently, large language models represented by ChatGPT[1] have garnered significant attention from researchers across various fields due to their exceptional ability to follow instructions and understand human intent, demonstrating outstanding performance in various natural language processing tasks.

Yi Ren, HanZhi Zhang, Jie He are with the Lab. of AI, Hangzhou Institute of Technology of Xidian University, Hangzhou, 311231, China. Tianyi Zhang, Weibin Li, Licheng Jiao are with the School of Artificial Intelligence, Xidian University, Xi'an, 710071, China. Jun Fu is with Xijing Hospital, The Fourth Military Medical University, Xi'an 710032, China. Diandong Liu are with the Shaanxi Joint Laboratory of Artificial Intelligence (Shaanxi University of Science and Technology), Xi'an 710021, China. (email: weibinli@xidian.edu.cn;)

Although the ChatGPT series remains closed-source, many excellent open-source large language models have emerged in the community, such as LLaMA[2], Mistral[3], and Qwen[4]. With the flourishing development of the open-source community, researchers have begun exploring the application of large language models in specific fields, such as healthcare. However, these professional fields often require highly specialized knowledge and terminology, and large language models typically lack such highly specialized expertise.

In the hydrology field, Ren et al. introduced WaterGPT[5], which, based on Qwen-7B-Chat and Qwen2-7B-Chat, underwent large-scale secondary pretraining and instruction tuning on domain-specific data, enabling professional knowledge Q&A and intelligent tool usage. In the field of remote sensing[6,7,8], Kuckreja et al. proposed GeoChat[9], the first multimodal large model[10] capable of understanding various types of remote sensing images. In the medical field, researchers have made significant progress in effectively utilizing large language models. For example, Google and DeepMind's research team published Med-PaLM[11] in Nature, with responses comparable to human clinicians 92.9% of the time. Additionally, Yunxiang Li et al. proposed ChatDoctor[12], which significantly improved the model's ability to understand patient needs, provide reasonable suggestions, and assist in various medical fields by fine-tuning the LLaMA model with a dataset of 205k doctor-patient dialogues. Haochun Wang et al. developed BenTsao[13] by instruction fine-tuning the LLaMA-7B model with a large-scale Chinese medical dataset, achieving remarkable results. Furthermore, Ming Xu et al. introduced MedicalGPT[14], which achieved significant advancements through the entire process from incremental pre-training to DPO[15] in applying large language models in healthcare.

Beyond textual research, many researchers are also exploring the integration of image and voice information. For instance, Rongsheng Wan et al.'s XrayGLM[16] processed large-scale bilingual medical multimodal datasets on MIMIC-CXR[17] and OpenI[18], fine-tuning the VisualGLM-6B model. Chunyuan Li et al. introduced LLaVA-Med[19], a multimodal large model specifically designed for biomedicine, while Chaoyi Wu et al. developed RadFM[20] by constructing a large-scale medical multimodal dataset MedMD, containing 16 million 2D or 3D images, and training a new medical multimodal large model RadFM based on this dataset.

Despite significant research achievements, actual medical diagnosis often requires integrating multiple modalities to achieve accurate results. To this end, we propose a new framework—MMDS, a system capable of recognizing medical images and patient facial details, and providing professional medical diagnoses. The system consists of two core components:

The first component is the analysis of medical images and videos. We specifically trained a multimodal medical model that can analyze medical images and accurately assess patients' facial emotions and facial paralysis conditions. This model achieved an accuracy of 72.59% on the FER2013[21] facial emotion recognition dataset, with a 91.1% accuracy in recognizing the "happy" emotion. In the task of facial paralysis recognition, the model achieved an accuracy of 93%, which is a 30% improvement compared to GPT-4o. Based on this model, we developed a parser for analyzing facial movement videos of facial paralysis patients, successfully achieving precise grading of the severity of facial paralysis. In tests on 30 videos of facial paralysis patients, the grading accuracy reached 83.3%.

The second component is generating professional medical responses. We employed a large language model integrated with a medical knowledge base to generate professional diagnostic opinions based on the analysis of medical images or videos. The core of this component is the construction of a departmental knowledge base routing mechanism based on the large language model. The model categorizes data by medical departments, and during the retrieval process, it decides which departmental knowledge base to query, significantly improving the retrieval accuracy in the RAG[22] (retrieval-augmented generation) process. This mechanism resulted in an average accuracy improvement of 4 percentage points for various large language models on the MedQA[23] dataset, and it achieved the highest accuracy of 84.41% for a 7B-level model on the MedQA Chinese evaluation dataset.The main contributions of this paper are as follows:

1.We propose an innovative Multimodal Medical Diagnosis Framework[24,25,26], which integrates the interpretation of medical images and videos with facial detail analysis. This framework enhances comprehensive diagnostic capabilities by combining multimodal analysis with specialized departmental knowledge, improving both the accuracy and efficiency of medical consultations.

2.We developed a specialized multimodal medical large model, achieving 93% accuracy in facial palsy recognition. It also reached 72.59% in emotion recognition, demonstrating the ability to perform various tasks related to facial detail analysis.

3.We created the first parser for analyzing facial movement videos of facial palsy patients, capable of accurately grading the severity of facial paralysis based on video analysis, with a grading accuracy of 83.3% in tests on 30 patients.

4.We introduced a novel departmental knowledge base routing management system within a RAG (retrieval-augmented generation) framework, significantly improving retrieval accuracy. This system achieved the highest accuracy of 84.41% for a 7B-level model on the MedQA Chinese evaluation dataset.

## II. DATASET CREATION

This study aims to explore how large language models can be better applied in the medical field. In actual medical scenarios, multiple modalities of information sources are often integrated, and conclusions are drawn based on analysis and reasoning of past case experiences. Therefore, to simulate the real diagnostic process of doctors, our research focuses on the construction of medical multimodal large models and the development of a hierarchical knowledge base mechanism.

### A Training Data

Our training data for the medical multimodal large model comprises the following components:

(1)LLaVA-Med[27]

We utilized the llava_med_instruct_60k dataset from the LLaVA-Med project as part of our training data. This dataset is based on biomedical image and text pairs from the PubMed Central[28] (PMC) database. It has been rigorously screened and processed, containing 60,000 high-quality image-text pairs.

(2)XrayGLM

We selected the OpenI-zh dataset from the XrayGLM project, which was created by preprocessing the chest X-ray dataset from Indiana University Hospital. This dataset includes 6,423 chest medical images and their corresponding Chinese and English diagnostic reports.

(3)Facial detail capture dataset[29]

To enhance the model's ability to capture facial details of patients, we constructed a facial dataset from multiple sources. First, we selected 500 facial images of Asian individuals from the CASIA-Face dataset, 500 facial images of normal individuals from the CelebA-Dialog dataset[30], and 1,000 facial images of facial palsy patients from the Facial Nerve Palsy Database[31]. These images were initially annotated using GPT-4o, followed by detailed annotation and data cleaning by specialists, resulting in our Facial Detail Capture Dataset.

To improve the model's accuracy in facial emotion recognition, we trained it using the training set of FER2013. This dataset is divided into seven emotion categories: sadness, anger, surprise, fear, happiness, disgust, and neutral, containing a total of 28,709 data samples.

By integrating the above datasets, we constructed a training dataset containing 97,132 image-text pairs, which was used to train our medical multimodal large model.

### B Multi-department Knowledge Base Dataset

To enable the large language model to function as specialized doctors from different departments for professional medical Q&A, we collected up to 1GB of high-quality doctor-patient dialogue data from the internet. We then used the large language model to precisely categorize the data according to different medical departments, ultimately constructing our multi-department knowledge base dataset.

*C Evaluation Datasets*

To validate the performance of our system, we used the following evaluation datasets:

1. Facial Palsy Evaluation Dataset: This dataset consists of 50 facial images from different facial palsy patients and 150 facial images of different normal individuals displaying various emotions such as Anger, Contempt, Disgust, Fear, Happy, Neutral, Sad, and Surprised. Since intense emotions can easily cause facial muscle distortion, this dataset effectively tests the model's ability to distinguish between facial palsy patients and normal individuals.

2. Facial Palsy Video Evaluation Dataset: Due to the difficulty of collecting videos of facial palsy patients, we only managed to gather 30 video datasets from online sources. These videos primarily feature patients performing actions such as laughing, smiling, frowning, puckering, showing lower teeth, and closing eyes, with patients typically evaluated at levels II to III on the House-Brackmann[32] (H-B) scale (As shown in Table 1). This dataset is used to verify whether our system can accurately grade the severity of facial palsy based on patient videos.

Table 1. House-Brackmann (H-B) scale

| Grade | Description | Characteristics |
|---|---|---|
| I | Normal Function | Normal facial function in all areas |
| II | Mild Dysfunction | Slight weakness noticeable only on close inspection, may have very slight synkinesis |
| III | Moderate Dysfunction | Obvious but not disfiguring difference between the two sides, noticeable but not severe synkinesis |
| IV | Moderately Severe Dysfunction | Obvious weakness and/or disfiguring asymmetry, severe synkinesis, incomplete eye closure |
| V | Severe Dysfunction | Barely perceptible movement, asymmetry at rest, no effective movement of the face |
| VI | Total Paralysis | No movement at all, loss of tone, asymmetry at rest |

3. FER2013: FER2013 is a dataset containing 35,887 grayscale facial images at a resolution of 48x48 pixels, labeled with seven emotion categories: anger, disgust, fear, happiness, sadness, surprise, and neutral. The dataset suffers from class imbalance, and emotion classification is challenging due to variations in facial pose, lighting, and age. The dataset is divided into a training set, a public test set, and a private test set, and is primarily used for facial expression recognition tasks. We used 7,178 images from the test set to evaluate our model's performance on the facial expression recognition task.

4. MedQA: MedQA is a large-scale open-domain question-answering dataset for the medical field, mainly composed of multiple-choice questions from professional medical qualification exams (such as the United States Medical Licensing Examination (USMLE) and other multiple-choice questions from mainland China and Taiwan. The dataset covers three languages: English, Simplified Chinese, and Traditional Chinese, with 12,723, 34,251, and 14,123 questions respectively. MedQA requires high levels of logical reasoning and prior knowledge integration, including many complex questions that need multi-hop reasoning. To provide the necessary background information for answering these questions, the dataset also includes extensive medical textbook text. We selected the test dataset with 3,426 multiple-choice questions applicable to mainland China to evaluate the performance of our system.

III. PROPOSED METHODS

*A MMDS system architecture*

This section describes the complete architecture of the Multimodal Hierarchical Department Consultation System, which consists of two stages:

**First Stage:** When a user inputs medical images or patient videos, these inputs are processed by a medical image parser and patient video parser built around the core medical multimodal large model. This stage involves a detailed analysis of the user's facial and body conditions as well as the medical images. The detailed analysis results are then passed on to the second stage.

**Second Stage:** The Medical Long Agent receives and summarizes the analysis results from the first stage. The Medical Long Agent is powered by a local large language model combined with specific prompt templates. Its responsibilities include:

1. Analyzing the medical image and video analysis results and the user's queries.

2. Determining whether further detailed descriptions from the user are needed or guiding the medical multimodal large model to analyze additional medical image information.

3. Identifying the appropriate department knowledge base to be used.

4. Retrieving relevant past informationp[33] from the department's knowledge base based on the user's symptoms and queries.

5. Acting as a specialist doctor from the relevant department, providing comprehensive answers and generating professional medical reports by combining case information, user symptoms, and user queries.

The medical report generated will be stored on blockchain nodes, sealed as historical information. When the user visits and asks questions again, the medical large model will

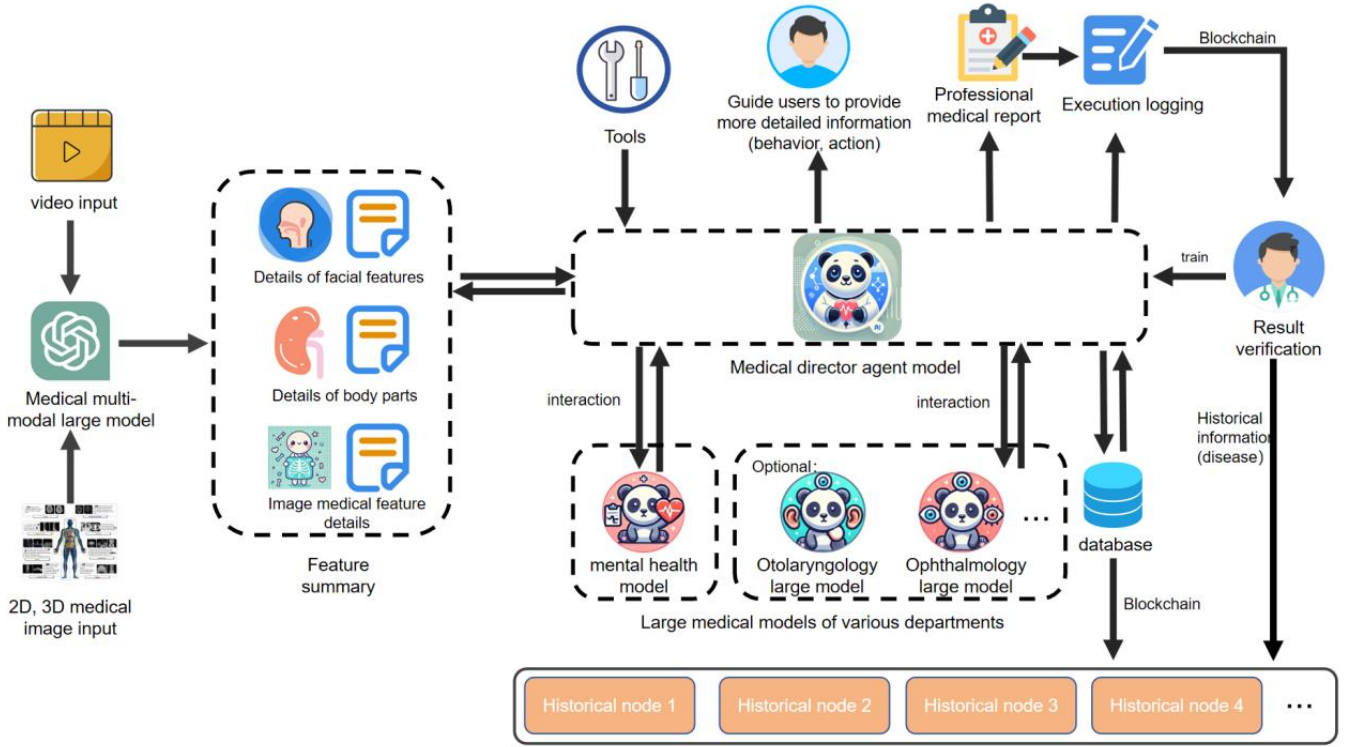

Fig1. MMDS system architecture diagram

combine historical information to provide a more professional medical report. The entire process will be recorded in an execution log, allowing doctors to intervene and verify at any time.

The complete architecture of the system is illustrated in Figure 1

### B MMDS Analysis of Medical Images

The core of the medical image parser is the medical multimodal large model. This model was fine-tuned on the InternLM-XComposer2-VL[34] model using LoRA[35] (Low-Rank Adaptation) training on a single A100 GPU, based on our collected training data for medical multimodal large models. The training parameters are listed in Table 2.

This model has the capability to analyze medical images, analyze users' facial emotions, and interpret facial images of patients to identify the presence of facial palsy.

Table 2. Train parameters

| Hyper parameter | Value |
| --- | --- |
| Precision | fp16 |
| Epochs | 6 |
| Max length | 4096 |
| Batch size | 8 |
| Weight_decay | 0.1 |
| Warmup_ratio | 0.01 |
| Learning rate | 5e$^{-5}$ |

### C MMDS Analysis of Medical Videos

Figure 2 shows the detailed process of our medical video parser, using the example of analyzing a video uploaded by a facial palsy patient. Our medical video parser consists of five modules: (i) multimodal preprocessing, (ii) external data collection, (iii) second-level frame video description generation, (iv) generation of complete video description script, and (v) generation of professional medical report. Each module is described in detail below.

(i) Multimodal Preprocessing[36]: Starting with the input video file, the video parser automatically uses ASR[37,38] tools to transcribe the speech in the video into text. Then, the video parser extracts 1 to 2 frames per second from the input video, which are then analyzed by the medical multimodal large model.

(ii) External Data Collection: For the incoming video, we collect the user's historical information, including previous conversation and symptom data, as well as some descriptive information about the video as external knowledge sources. This integrated information is included in the input prompts for both the analysis of images by the medical multimodal large model and the final generation of the complete video description by the medical large language model.

(iii) Second-level Frame Video Description Generation: For each second-level image, we designed specific prompt templates and used multiple queries to analyze each image's information.

(iv) Generation of Complete Video Description Script: At this stage, the video parser uses the medical large language

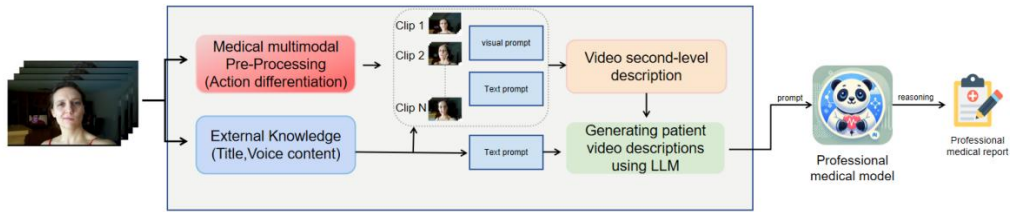

Fig2. MMDS analysis process for medical videos model to integrate detailed analyses of each image, the corresponding time information, and some external video description data to generate a comprehensive summary of the user's video. (v) Generation of Professional Medical Report: At this stage, the video parser uses the medical large language model to generate a detailed medical report by integrating the user's video description[39], user requests, historical information, and specific prompt templates. Additionally, these second-level video descriptions are integrated into a video script for easy query and verification in the future.

## D Multi-Department Knowledge Base Routing Management Mechanism Driven by Large Models

This section introduces another core component of the MMDS system: a multi-department knowledge base routing management mechanism driven by large models.

The process begins with using a large language model to generate a medical question-and-answer dataset based on medical text data. This dataset is then classified according to a predefined list of medical departments. Once classified, the data is stored in knowledge bases named after each corresponding department. After analyzing medical images and videos, the analysis results are sent to our locally deployed large language model. The model first determines whether the current data is sufficient for making a decision or if more detailed information is needed. Once the necessary data is collected, the model autonomously identifies which department's knowledge base should be consulted.

Once the appropriate department is identified, our system automatically retrieves relevant medical cases from the corresponding knowledge base. The local large model then acts as the department's doctor, reading through the relevant medical cases and the user's detailed data. Step by step, the model produces a comprehensive medical report.

The process flow is illustrated in Figure 3, and the algorithm workflow is depicted in Figure 4.

In this system, we fine-tuned the bge-large-zh-v1.5[40] and BERT[41] models based on the collected multi-department knowledge base dataset. These models serve as the medical recall and rank models, respectively. They are used to finely retrieve the necessary cases based on the user's queries.

## IV. EXPERIMENTS AND ANALYSIS

### A Emotion Recognition Task

We evaluated the accuracy of our medical large model on the FER2013 test dataset, which contains a total of 7,178 images. The confusion matrix results are shown in Figure 5, where A, B, C, D, E, F, and G represent the emotion categories of sadness, anger, surprise, fear, happiness, disgust, and neutral, respectively. Additionally, we compared various classic models with our model on the emotion recognition task, and the results are shown in Table 3.

**Table 3.** The confusion matrix of MMDS in recognizing various emotions.

| Method | Accuracy Rate |
|---|---|
| CNN | 62.44 |
| GoogleNet[42] | 65.20 |
| VGG+SVM | 66.31 |
| Conv+Inception layer | 66.40 |
| Bag of Words [43] | 67.40 |
| Attentional ConvNet 44] | 70.02 |
| CNN + SVM | 71.20 |
| ARM (ResNet-18)[45] | 71.38 |
| Inception [46] | 71.60 |
| ResNet [47] | 72.40 |
| **MMDS** | **72.59** |

In this experiment, we compared the performance of different methods on the emotion recognition task. As shown in Table 3, we tested a variety of models, including CNN, GoogleNet, VGG+SVM, Conv+Inception layer, Bag of Words, Attentional ConvNet, CNN+SVM, ARM (ResNet-18), Inception, ResNet, and MMDS. Among these models, MMDS achieved the highest accuracy rate, reaching 72.59%.

Specifically, the traditional Convolutional Neural Network (CNN) method achieved an accuracy of 62.44% on the emotion recognition task, demonstrating relatively weak performance. As the network structure became more complex, the performance of GoogleNet and VGG+SVM improved, reaching 65.20% and 66.31%, respectively. The Conv model with an added Inception layer further improved the accuracy to 66.40%, indicating that modular designs in deep networks (such as the Inception layer) can better capture emotion-related features.

In more advanced models, the Bag of Words and Attentional ConvNet models achieved accuracy rates of 67.40% and 70.02%, respectively. Notably, the Attentional

```
Algorithm 1 Medical Text Processing and Knowledge Base Retrieval Proce-
dure
 Input: Original medical text data MT, List of medical departments
 Departments = {D_1, D_2, ..., D_n}, Pretrained language model LM, User
 query Q
 Output: Answer to the user query or appropriate feedback
 Step 1: Generate Question-Answer (QA) Dataset
 QA_dataset ← LM(MT)    Use language model to generate QA pairs based on
 MT
 Step 2: Classify QA Pairs into Departments
 for each qa ∈ QA_dataset do
   dept ← LM(qa, Departments)    Classify the QA pair into the appropriate
 department
   Store qa into the knowledge base of dept Save the classified QA pair to the
 corresponding department's knowledge base
 end for
 Step 3: Handle User Query
 query_dept ← LM(Q, Departments)    Identify the relevant department based
 on user query Q
 response ← Retrieve answer from knowledge base of query_dept    Fetch the
 most relevant answer from the department's knowledge base
 if response exists then
   return response                    Return the answer to the user
 else
   return "Sorry, we could not find a suitable answer. Please provide more
 details."                    Fallback if no relevant answer is found
 end if
```

Fig3.Process Flow of the Multi-Department Knowledge Base Routing Management Mechanism

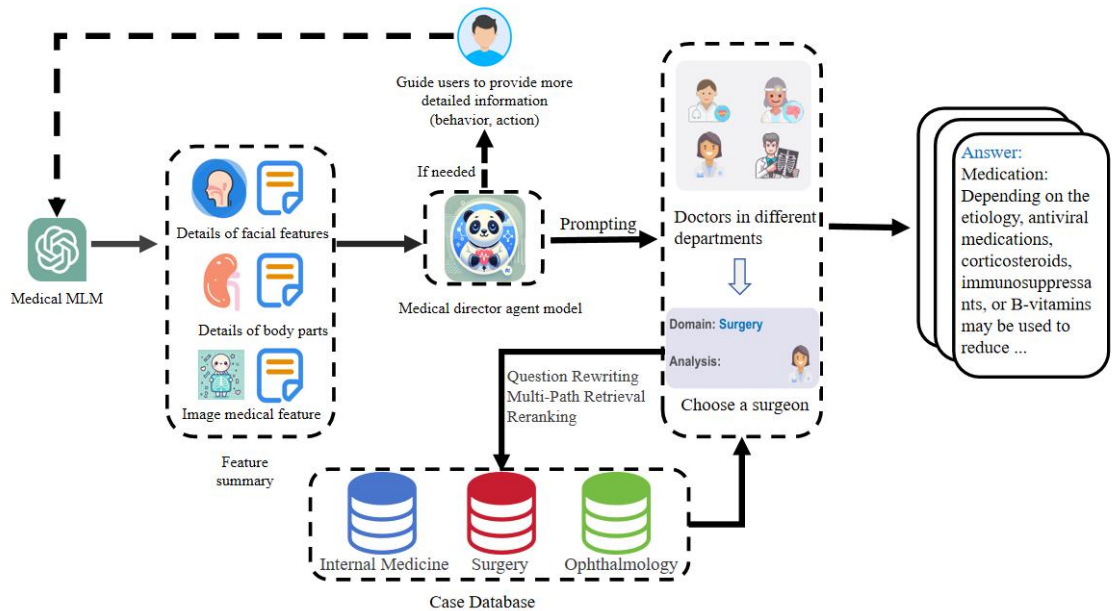

Fig4.Multi-department knowledge base management mechanism architecture diagram in MMDS

ConvNet introduced an attention mechanism that focuses on key feature information, significantly improving recognition performance. Meanwhile, models that combined Support Vector Machines (SVM) with deep neural networks also performed well, with CNN+SVM achieving an accuracy of 71.20%.

In deeper network models, ARM (ResNet-18) and ResNet achieved accuracy rates of 71.38% and 72.40%, respectively. In particular, the deeper ResNet model effectively avoided gradient vanishing issues, leading to more stable and precise performance in complex emotion recognition tasks. The Inception model achieved an accuracy of 71.60%, further demonstrating the effectiveness of modular network structures. Finally, the proposed MMDS system, which includes a large multimodal model, achieved the highest accuracy of 72.59% in this experiment, surpassing all other compared models. The success of the MMDS model can be attributed to its multimodal fusion design, which integrates different feature types, demonstrating exceptional generalization capabilities and precision in emotion recognition tasks.

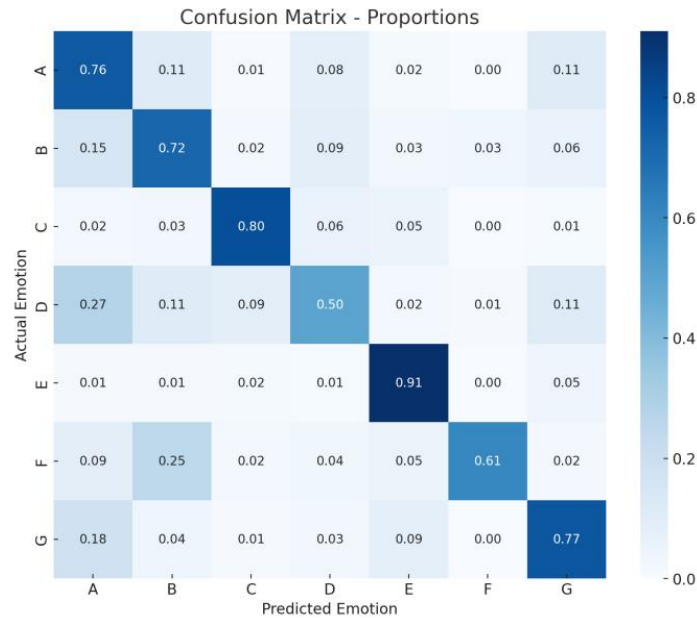

Fig5. The confusion matrix of MMDS in recognizing various emotions.

To further analyze the classification performance of each emotion category, we plotted the confusion matrix (see Figure 5). From the confusion matrix, we can see that MM performed best in recognizing the emotion categories of "happiness" and "surprise," achieving accuracy rates of 91.1% and 80.27%, respectively. This indicates that the MMDS model is highly sensitive and accurate in recognizing emotions with high intensity.

However, the model's performance was slightly weaker in recognizing emotions like "fear" and "anger," with accuracy rates of 49.62% and 72.095%, respectively. This may be due to the complexity of these emotions' expressions in real-world scenarios, making them more prone to confusion with other emotions. In particular, in the classification of "fear," the confusion matrix shows a high misclassification rate, with many "fear" samples being misclassified as "sadness" or "neutral," further highlighting the challenges of feature extraction for this emotion category.

In addition, the model performed reasonably well in the "neutral" category, achieving an accuracy rate of 77.445%, indicating that MMDS can accurately capture features associated with non-emotional states. As for the "disgust" category, although the sample size for this category was relatively small, the model still achieved an accuracy of 61.33%, demonstrating its strong generalization capabilities.

Overall, the MMDS model achieved excellent performance in the emotion recognition task, especially in the "happiness" and "surprise" categories. Despite some limitations in recognizing the "fear" category, the MMDS model is capable of accurately capturing facial details and successfully completing multiple facial analysis tasks.

## B MMDS Facial Palsy Recognition Accuracy

We evaluated our medical multimodal large model, along with a batch of the most advanced multimodal large models, on our constructed facial palsy evaluation dataset. The results are shown in Table 4.

As seen in the table, among the untrained models, GPT-4o performed the best, with an overall accuracy of 63%. It had an accuracy of 88% in recognizing facial palsy patients but only 54.67% accuracy in identifying normal individuals, often mistakenly identifying normal individuals displaying different emotional expressions as having facial palsy. In contrast, our model achieved an overall accuracy of 92%, with errors occurring only when identifying emotional expressions in normal individuals, but still maintaining a 90.67% accuracy rate in this regard, surpassing GPT-4o by 30% overall.

**Table 4.** Recognition accuracy of images of patients with facial paralysis by each multi-modal large model

| Model | Facial Palsy Patients | Normal Individuals | Overall Accuracy |
|---|---|---|---|
| MMDS | 100 | 90.67 | 93 |
| Gpt4o | 88 | 54.67 | 63 |
| Gpt4o-mini | 80 | 26 | 39.5 |
| Glm4[48] | 96 | 6.67 | 29 |
| Internlm-Xcomposer2-Vl-7b | 20 | 66.67 | 55 |

These results indicate that our medical multimodal large model significantly outperforms other models in the facial palsy recognition task. This superior performance is primarily attributed to our more refined preprocessing and annotation .

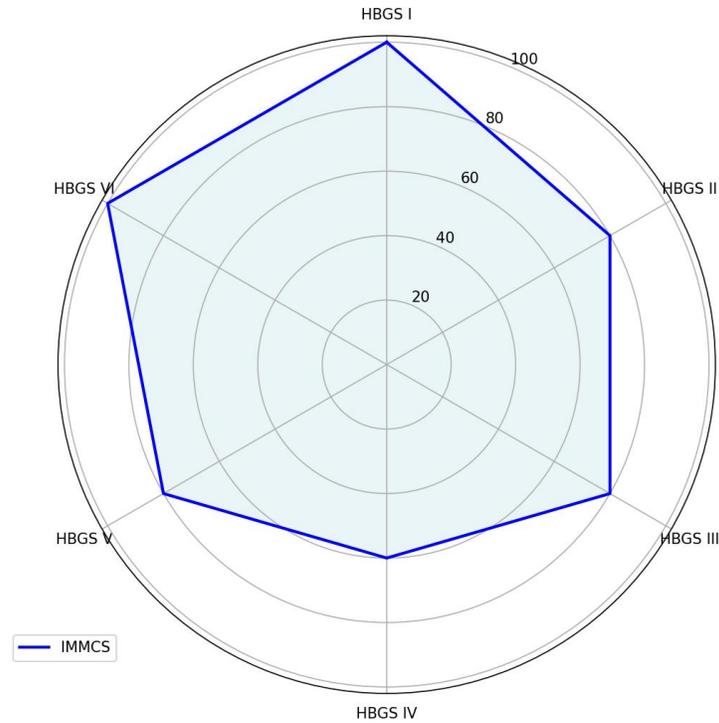

**Fig6.** MMDS patient upload video analysis process of the data during the training process and the enhanced training for understanding images of both normal individuals and facial palsy patients

## C MMDS Accurate Grading of Facial Palsy Based on Patient Videos

In this section, we will detail the architecture and experimental results of our medical video parser. For user-uploaded videos, the medical video parser extracts frames per second and then analyzes each image to determine whether the patient has facial palsy. If more than half of the frames indicate facial palsy, the system proceeds with a detailed analysis of the video. The parsing process for videos of facial palsy patients is shown in Figure 6.

To further validate the performance of our system, we tested it using videos collected from 30 patients with facial paralysis. The overall accuracy of our system on the test dataset was 83.33%, with the accuracy for each level shown in Figure 7.

As illustrated in the figure, the accuracy for HBGS I and HBGS VI levels was 100%. However, there were errors in recognizing levels II to V, with level IV showing an accuracy of only 60%. This indicates that video descriptions for levels II to V have many similarities, making it challenging to distinguish between these levels accurately. Detailed and comprehensive video descriptions are required for better differentiation. Using video descriptions for grading facial palsy patients is more logical, explanatory, and provides a more intuitive basis for medical diagnosis and recommendations.

## D Accuracy of Multi-department Knowledge Base Management Mechanism on MedQA

In this study, we evaluated a series of state-of-the-art large language models using the mainland China dataset from MedQA to test the graded management mechanism of medical knowledge bases. These models include various models from different research institutions and companies, covering both API and weight access methods. The evaluation results are shown in Table 5.

As can be seen from the table, the performance of the Qwen2-7B-instruct model improved from 80.64 to 84.41 after introducing the classified RAG mechanism, demonstrating the effectiveness of this mechanism in complex knowledge base scenarios. Additionally, the InternLM2-Chat-7B model also showed improvement after introducing the classified RAG, with its score increasing from 53.47 to 57.78. Compared to closed-source models accessed via API, locally managed models under the graded knowledge base mechanism showed comparable accuracy. For example, the accuracy of GLM-4 reached 84.73, only 0.32 higher than the local model, while the best closed-source model, GPT-4O, achieved an accuracy of 86.05, only 1.64 higher than the local model. The accuracy of DeepSeekv2 was even 1.71 lower than the local model. However, when using the traditional RAG mechanism without graded management, the performance of the Qwen2-7B-instruct [49] model decreased from 80.64 to 76.07.

Table 5. Evaluation results of each model on MedQA (Mainland)

| Model | Creater | #Parameters | Access | MedQA (Mainland) |
|---|---|---|---|---|
| GPT-4o-mini | OpenAI | undisclosed | API | 75.37 |
| GPT-4o | OpenAI | undisclosed | API | 86.05 |
| DeepSeek-V2[50] | DeepSeek-AI | 236B | API | 82.70 |
| GLM-4 | Zhipu AI | undisclosed | API | 84.73 |
| GLM-4-9B-chat | Zhipu AI | 9B | Weights | 80.93 |
| InternLM2-Chat-7B | Shanghai AI Lab | 7B | Weights | 53.47 |
| InternLM2-Chat-7B-classification-RAG | Shanghai AI Lab | 7B | Weights | 57.78 |
| Qwen2-7B-instruct | Alibaba Group | 7B | Weights | 80.64 |
| Qwen2-7B-instruct+rag | Alibaba Group | 7B | Weights | 76.07 |
| Qwen2-7B-instruct+classification_rag | Alibaba Group | 7B | Weights | 84.41 |

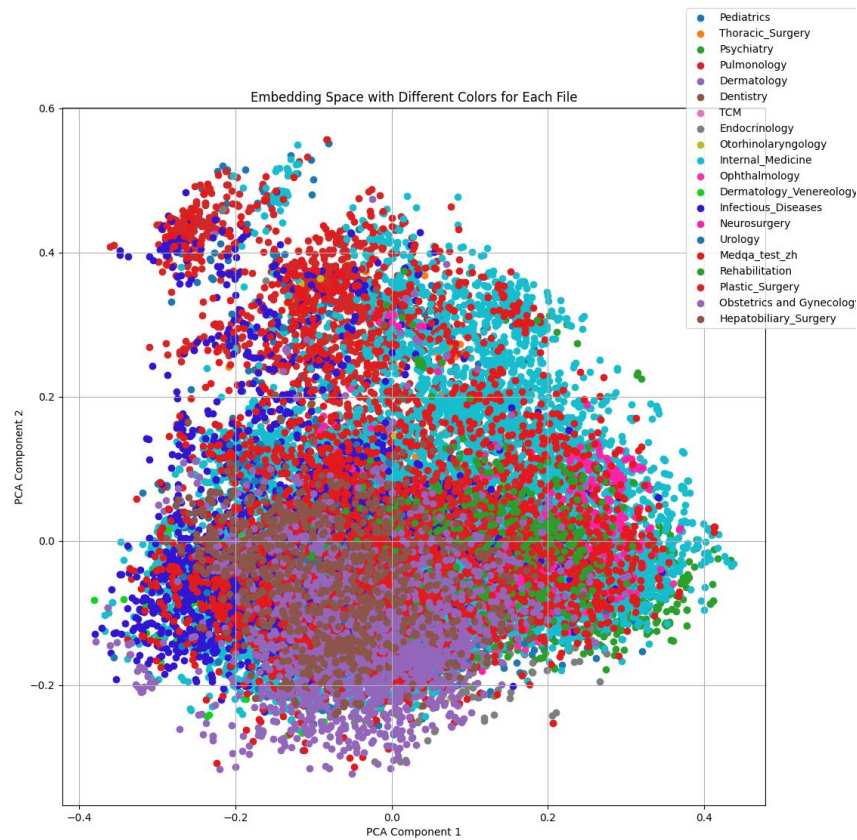

**Fig8**. Vector space projection diagram of knowledge base data of each department and MedQA (Mainland) test data

To analyze the reasons behind this result, we projected the knowledge base data of each department and the test data from MedQA_zh into the vector space using our trained medical embedding model. The results are shown in Figure 8.

In the figure, the red represents the test data, while the other colors represent the datasets from different departments. As observed, there is a significant overlap among many departmental datasets, indicating semantic similarity in the vector space. Given that our knowledge base dataset is as large as 1GB, this overlap can easily lead to substantial data conflicts. This is why, under the traditional RAG mechanism, it is difficult to recall high-quality data. Conversely, the graded management of the knowledge base effectively alleviates data conflicts and enables the recall of relatively

2high-quality data, thereby significantly improving the quality of the model's responses.As seen in the figure, our model equipped with the department-managed knowledge base achieves higher scores in almost every department compared to the baseline models, further proving the advantage of our mechanism.

## V. CONCLUSION

In this paper, we introduced MMDS, a comprehensive system capable of recognizing medical images and patient facial details to provide professional medical diagnoses. The system comprises two core components: a specialized multimodal medical model for analyzing medical images and videos, and a large language model integrated with a medical knowledge base for generating expert medical responses.

Our multimodal medical model demonstrated high accuracy in facial emotion recognition, achieving a 72.59% accuracy rate on the FER2013 dataset and excelling particularly in recognizing the "happy" emotion with a 91.1% accuracy. In facial paralysis recognition, the model reached an impressive 92% accuracy, outperforming GPT-4o by 30%. By developing a parser to analyze facial movement videos of patients with facial paralysis, we achieved precise grading of paralysis severity, with the system demonstrating an 83.3% grading accuracy in tests on 30 patient videos.

The second component of MMDS leverages a large language model augmented with a department-specific knowledge base routing management mechanism. This innovation allows the model to categorize data by medical departments and determine the appropriate knowledge base during retrieval, significantly improving retrieval accuracy in the RAG (retrieval-augmented generation) process. This mechanism led to an average increase of 4 percentage points in accuracy for various large language models on the MedQA dataset.

The results highlight MMDS's potential as a valuable tool in medical diagnostics, offering accurate image analysis and professional diagnostic capabilities. The system's ability to integrate multimodal data and generate expert-level responses can assist healthcare professionals in making informed decisions. Future work will focus on expanding the system's capabilities to cover a broader range of medical conditions and further refining the knowledge base routing mechanism to enhance diagnostic accuracy.

Our code is open-sourced and available at https://github.com/renllll/MMDS. We encourage the research community to utilize and build upon our work to advance the field of medical image analysis and automated diagnosis.

**Funding:** This research was supported by Key Projects of Shaanxi Provincial Department City Cooperation (2022GD-TSLD-61-3), Xi'an Science and Technology Plan Project(23ZDCYTSGG0026-2022, 24LLRHZDZX0016), Shhanxi Coal Geology Group Co. Ltd Scientific Research Project (SMDZ-2023CX-14)## REFERNCES

[1] Ouyang L, Wu J, Jiang X, et al. Training language models to follow instructions with human feedback[J]. Advances in neural information processing systems, 2022, 35: 27730-27744.
[2] Dubey A, Jauhri A, Pandey A, et al. The llama 3 herd of models[J]. arXiv preprint arXiv:2407.21783, 2024.
[3] Hou G, Lian Q. Benchmarking of Commercial Large Language Models: ChatGPT, Mistral, and Llama[J]. 2024.
[4] Bai J, Bai S, Chu Y, et al. Qwen technical report[J]. arXiv preprint arXiv:2309.16609, 2023.
[5] Y. Ren, T. Zhang, X. Dong, et al., "WaterGPT: Training a large language model to become a hydrology expert," SSRN, no. 4863665, 2024.
[6] Zhang, T., Li, W., Feng, X., et al. (2024). Super-resolution water body extraction based on MF-SegFormer. In IGARSS 2024-2024 IEEE International Geoscience and Remote Sensing Symposium (pp. 9848-9852). IEEE.
[7] Liu, F., Chen, D., Guan, Z., et al. (2024). RemoteCLIP: A vision language foundation model for remote sensing. IEEE Transactions on Geoscience and Remote Sensing.
[8] Zhang, Z., Zhao, T., Guo, Y., et al. (2023). RS5M: A large scale vision-language dataset for remote sensing vision-language foundation model. arXiv preprint arXiv:2306.11300.
[9] Kuckreja K, Danish M S, Naseer M, et al. Geochat: Grounded large vision-language model for remote sensing[C]//Proceedings of the IEEE/CVF Conference on Computer Vision and Pattern Recognition. 2024: 27831-27840.
[10] Y. Ren, T. Zhang, Z. Han, et al., "A Novel Adaptive Fine-Tuning Algorithm for Multimodal Models: Self-Optimizing Classification and Selection of High-Quality Datasets in Remote Sensing," arXiv preprint arXiv:2409.13345, 2024.
[11] Singhal, K. et al. "Large language models encode clinical knowledge." Nature 620 (2022): 172 - 180.
[12] Li, Yunxiang et al. "ChatDoctor: A Medical Chat Model Fine-Tuned on a Large Language Model Meta-AI (LLaMA) Using Medical Domain Knowledge." Cureus 15 (2023): n. pag.
[13] Wang, Hao et al. "HuaTuo: Tuning LLaMA Model with Chinese Medical Knowledge." ArXiv abs/2304.06975 (2023): n. pag.
[14] Kraljevic, Zeljko et al. "MedGPT: Medical Concept Prediction from Clinical Narratives." ArXiv abs/2107.03134 (2021): n. pag.
[15] Rafailov, Rafael et al. "Direct Preference Optimization: Your Language Model is Secretly a Reward Model." ArXiv abs/2305.18290 (2023): n. pag.
[16] Rongsheng Wang, Yaofei Duan, Junrong Li, Patrick Pang, & Tao Tan. XrayGLM [Computer software]. https://github.com/WangRongsheng/XrayGLM
[17] Johnson, Alistair E. W. et al. "MIMIC-CXR-JPG, a large publicly available database of labeled chest radiographs." (2019).
[18] Demner-Fushman, Dina et al. "Design and Development of a Multimodal Biomedical Information Retrieval System." J. Comput. Sci. Eng. 6 (2012): 168-177.
[19] Li, Chunyuan et al. "LLaVA-Med: Training a Large Language-and-Vision Assistant for Biomedicine in One Day." ArXiv abs/2306.00890 (2023): n. pag.
[20] Wu, Chaoyi et al. "Towards Generalist Foundation Model for Radiology." ArXiv abs/2308.02463 (2023): n. pag.
[21] Goodfellow I J, Erhan D, Carrier P L, et al. Challenges in representation learning: A report on three machine learning contests[C]//Neural information processing: 20th international conference, ICONIP 2013, daegu, korea, november 3-7, 2013.